\documentclass[conference]{IEEEtran}
\IEEEoverridecommandlockouts
\usepackage{cite}
\usepackage{amsmath,amssymb,amsfonts}
\usepackage{algorithmic}
\usepackage{graphicx}
\usepackage{textcomp}
\usepackage{xcolor}

\usepackage{subcaption}
\usepackage[normalem]{ulem}

\usepackage{verbatim}
\usepackage[ruled,linesnumbered]{algorithm2e}
\makeatletter
\newcommand{\algorithmicforeach}{\textbf{for each}}
\newcommand{\FOREACH}[2][default]{%
  \ALC@it\algorithmicforeach\ #2\ \algorithmicdo%
    \ALC@com{#1}\begin{ALC@for}}
\makeatother

\usepackage{xcolor}

\usepackage[utf8]{inputenc}
\usepackage{fourier} 
\usepackage{array}
\usepackage{makecell}

\def\BibTeX{{\rm B\kern-.05em{\sc i\kern-.025em b}\kern-.08em
    T\kern-.1667em\lower.7ex\hbox{E}\kern-.125emX}}

\begin{document}

\title{Deep Learning Based Face Recognition Method using Siamese Network}

\author{\IEEEauthorblockN{Enoch Solomon}
\IEEEauthorblockA{\textit{Department of Computer Science} \\
Virginia State University \\
Richmond, Virginia \\
esolomon@vsu.edu
}

\and
\IEEEauthorblockN{Abraham Woubie}
\IEEEauthorblockA{\textit{Silo AI} \\
Helsinki, Finland\\
Abraham.zewoudie@silo.ai}

\and

\IEEEauthorblockN{Eyael Solomon Emiru}
\IEEEauthorblockA{\textit{Department of Information} \\
Engineering and Computer Science \\
University of Trento, Italy \\
eyael.emiru@studenti.unitn.it}
}

\maketitle

\begin{abstract}

Achieving state-of-the-art results in face verification systems typically hinges on the availability of labeled face training data, a resource that often proves challenging to acquire in substantial quantities. In this research endeavor, we proposed employing Siamese networks for face recognition, eliminating the need for labeled face images. We achieve this by strategically leveraging negative samples alongside nearest neighbor counterparts, thereby establishing positive and negative pairs through an unsupervised methodology. The architectural framework adopts a VGG encoder, trained as a double branch siamese network. Our primary aim is to circumvent the necessity for labeled face image data, thus proposing the generation of training pairs in an entirely unsupervised manner. Positive training data are selected within a dataset based on their highest cosine similarity scores with a designated anchor, while negative training data are culled in a parallel fashion, though drawn from an alternate dataset. During training, the proposed siamese network conducts binary classification via cross-entropy loss. Subsequently, during the testing phase, we directly extract face verification scores from the network's output layer. Experimental results reveal that the proposed unsupervised system delivers a performance on par with a similar but fully supervised baseline.
\end{abstract}


\section{Introduction} 

In today's increasingly digital world, safeguarding personal information and ensuring security has become a top priority \cite{kumar2014survey,solomon2023fass,solomon2023hdlhc,cima,scope,ics_sea,memory_safety,fmemory_safety,earic,containers_security}. One of the most cutting-edge technologies addressing this concern is face verification, a powerful tool that leverages facial recognition to authenticate individuals \cite{kumar2009attribute}. This technology offers immense potential across various industries, from financial services to law enforcement, and even everyday consumer applications.

Deep learning methods are successful in image application specifically in face recognition \cite{serengil2020lightface, woubie2021federated, woubie2021voice,woubie2021federatedondevice,woubie2021use}. These approaches excel in learning deep features and bottleneck features (BNF) \cite{prasad2020deep}, which are used within a conventional GMM-UBM framework \cite{mallouh2018new}. However, these deep learning approaches typically rely on labeled training data.

The authors in \cite{ioffe2006probabilistic} reported that PLDA is the efficient back-end for image recognition \cite{el2013scalable,ioffe2006probabilistic}. Previous works reported PLDA provide a superior performance than cosine scoring. But, this improvement comes with the cost of labeled training data. However, previous works show that it is not usually easy to have access to large amounts of labeled training data. Thus, the lack of labeled training data results in a significant performance gap between cosine and PLDA scoring techniques \cite{hafed2001face,wang2018cosface} in face recognition. Although the authors in \cite{gao2017semi} proposed automatic labeling techniques, they could not appropriately estimate the true labels. Although these approaches perform reasonably well, the results are still compared to that PLDA that uses actual labels \cite{muslihah2020texture}. Whereas, the authors in \cite{hammouche2022gabor} proposed autoencoder based unsupervised method improve the performance of face recognition systems. Previous studies show that these approaches mainly aim at increasing the discriminative power of face image embeddings. Thus, they can be applied as a back-end in face verification task.

In our proposed work, the main goal is to reduce the reliance on labeled training data for a face verification system. We aim to obtain end-to-end face verification scores without using face image labels. We propose a siamese network \cite{chen2021exploring} consisting of double-branch networks, each with two branches that are CNN encoders. These encoders are inspired by the VGG architecture and adapted for face verification  \cite{cheng2020facial}. Traditionally, Siamese networks are trained using pairs of training samples, such as anchor-positive and anchor-negative pairs. However, to avoid using face image labels, we generate these training sample pairs in an unsupervised manner. Positive training data are chosen within one dataset based on their highest cosine scores with the anchor sample. In contrast, negative training data are selected in a similar manner from another dataset, ensuring that the two datasets do not include face images from the same identity.

In our study, we introduced the utilization of a two branch Siamese network. The network processes positive and negative samples, individually paired with anchor samples, and computes the binary cross-entropy loss based on their binary labels. Following the training process, decision scores for face verification trials are derived from the last layer of the network, resulting in an end-to-end face verification system. Our evaluation was performed on the Labeled Faces in the Wild (LFW) dataset \cite{huang2008labeled}. Despite being unsupervised, our results show significant promise, approaching the performance of a fully supervised baseline.

The remainder of this paper is organized as follows: Section 2 explains the proposed method, Section 3 provides details on the experimental setup and data, Section 4 discusses the results, and we conclude in Section 5.

\section{Proposed Method}

The training of a DNN in an end-to-end fashion typically requires access to labeled training data, which can be challenging in real-world scenarios. To address this, especially when labeled data is unavailable, we propose the use of siamese networks \cite{chen2021exploring}, characterized by a double-branch architecture. Each branch features a CNN encoder that is motivated by the VGG network \cite{sengupta2019going}, which has recently shown promise in face verification \cite{gwyn2021face}. Given our objective of avoiding reliance on face image labels, we introduce a novel approach to generate pairs of training data in an unsupervised manner \cite{solomon2022uface,solomon2023unsupervised,solomon2023autoencoder,woubie2023image,solomon2023face,https://doi.org/10.25772/re06-av14}. Specifically, positive training data are chosen within one dataset based on their similarly score with the anchor. Similarly, negative training data are selected using the same criterion, but from a separate dataset, ensuring that the two datasets do not contain face images from the same identity. This innovative training methodology circumvents the need for labeled data, making the process more accessible and adaptable to real-world scenarios.

The proposed work utilizes a siamese network. It is trained by minimizing binary cross-entropy loss. This network selects the pair of negative-anchor and  positive-anchor face images samples, and processes them to generate binary labels of 1 (positive match) or 0 (negative match) at the output. These labels are used to compute the loss during training. Once the network is trained, we can obtain decision scores for face recognition evaluations directly from the network's output. In summary, our proposed siamese network constitutes an end-to-end face recognition system, capable of making binary classification decisions with respect to facial similarity.

\begin{figure}[]
	\centering
 \centerline{\includegraphics[scale=0.35]{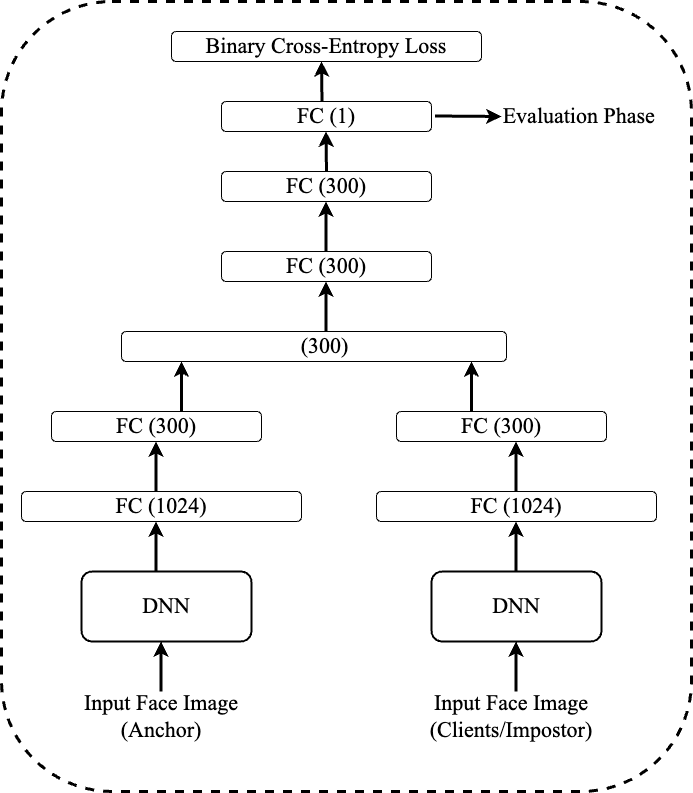}}
	\caption{The architecture of the proposed siamese network. FC denotes fully connected layer. In evaluation phase, decision scores for face recognition are obtained directly from the last layer of the network. This layer computes the final representation of the features extracted from the input face images, which is then used to make binary classification decisions regarding facial similarity.}
	\label{fig:arch2}
\end{figure}

  \begin{algorithm}
    \caption{The proposed algorithm to select positive and negative face images for each face image in the training dataset.}
    \label{algo}
    \begin{algorithmic}[]
      \REQUIRE Training face images $t_i$, 1 < i < n, $x_i \in X$ and $y_i \in Y$, $1 \leq i \leq n$, and threshold
      \ENSURE Positive and Negative face images $P_{i j}$ and $N_{i j}$,
$1 \leq i \leq m$ and $1 \leq j \leq k$
      \FOREACH{training face image $x_i$}
            \FOREACH{training face image  $x_p$, 1 < p < m }           
               \STATE Compute Positive Scores ${ }_{i,p}=\operatorname{cosine}\left(x_i, x_p\right)$                  
            \ENDFOR \\
       From Positive Scores ${ }_{i, p}$, select $k$ highest face images.
       
       \textbf{if} Positive Scores ${ }_{i, p}$ < threshold
        
            \qquad   $P_{i, j}=x_p$
        
            \FOREACH{training face image  $y_m$, 1 < n < m } 
                \STATE Compute Negative Scores ${ }_{i,n}=\operatorname{cosine}\left(y_i, y_n\right)$         
            \ENDFOR \\
       From Negative Scores ${ }_{i, n}$, select $k$ highest face images.
       
       \textbf{if} Negative Scores ${ }_{i, n}$ < threshold
            
            \qquad    $N_{i, j}=y_n$
      \ENDFOR
    \end{algorithmic}
  \end{algorithm}

\subsection{Training Pairs Selection}

The procedure for selecting negative and positive images involves utilizing two datasets, denoted as X and X. Let 
$C_X$ and $C_Y$ represent the individuals in datasets X and Y, respectively. It's assumed that individuals in dataset X are distinct from those in dataset Y, i.e., 
 $C_X \cap C_Y=\phi$. Algorithm \ref{algo} provides an overview of how the selection of negative and positive face image vectors is conducted in an unsupervised manner.

To begin, we extract the face embedding vectors for all face images in datasets X and Y. Next, we compute similarity scores among all face embedding vectors within dataset X by using cosine distance metrics. For each face embedding vector in X, we select a fixed number 
$k$ of neighbor face embedding as a potential positive face embeddings. Subsequently, we introduce a threshold to select potential $k$ positives face images. 

To select the negative face image embeddings, we score all the face embeddings in $X$ with those in $Y$, by using cosine similarity scores. For each face embedding in $X$, we select $k$ number of face embeddings from $Y$ that by using their cosine distance scores. Given that the person in dataset $X$ does not appear in dataset $Y$, the $k$ selected face embeddings are considered potential negative face images. Subsequently, we apply a threshold to their corresponding cosine scores to identify the most challenging negative face images. This process ensures that each face embedding in $X$ is paired with $k$ most positive and $k$ negative face image embeddings.

Assuming there are $n$ face images in each dataset $X$ and $Y$, then we will have a total of  n plus n times k face image training data. For example, if we have 100 training data in each datasets with k=10, then we will have (100+100)*10 = 1000 training data. This way, we were able to increase intrinsically the number training dataset. In our experiments, training pairs consist of two samples, namely anchor-positive and anchor-negative.

To select the negative face embedding vectors, we compute cosine similarity scores between all face embedding vectors in set $X$ and those in set $Y$. For each face embedding vector in $X$, we choose $K$ face embedding vectors from $Y$ that have the closest scores. Given that individuals in $X$ do not appear in $Y$, these $K$ selected face embeddings represent potential negative face images. Subsequently, we apply a threshold to the corresponding cosine scores to identify the most challenging negatives face images among them. As a result, each face embedding vector in $X$ is assigned $K$ positive and $K$ negative face embedding vectors.

\subsection{Proposed Siamese Network}

The training of a DNN in an end-to-end manner usually requires labeled face images for the training data, which can be challenging to obtain in practical scenarios. When such labels are unavailable, we propose the use of siamese networks. These networks operate in a fully unsupervised manner, distinguishing them from a typical DNN classifiers. Traditionally, a siamese network is trained using pairs of samples, consisting of an anchor, a positive, and negative. To bypass the need for face image labels, we propose generating these training pairs in an unsupervised manner, which are then fed into the network. Specifically, we advocate for a siamese network utilizing binary cross-entropy loss. Following the training process, we extract decision scores from the output of the proposed network for the evaluation. This approach allows us to perform face verification in an unsupervised manner.

\subsection{End-to-End Face Verification System}

Figure \ref{fig:arch2} depicts the architecture of the proposed siamese network. This network comprises two identical branches, both acting as CNN encoders. During training, a pair consisting of an anchor face image along with either a positive or negative face image is inputted into each branch of the network respectively. These two branches share the same parameters. Following the CNN encoder, the encoded vectors output from the two branches are   concatenated. Then, the concatenated vector is fed into the FC layers. The last layer is a binary class layer. It will be one for the case of anchor and positive face images, whereas it will be zero for the case of anchor and negative face images. Throughout the training process, the network minimizes binary cross-entropy loss.

After training with selected positive and negative samples in an unsupervised manner, evaluations are conducted in an end-to-end way. During evaluation, airs of test face images are inputted into the network, and decision scores are directly obtained from the last layer. This approach establishes an unsupervised end-to-end face verification system using the proposed Siamese network.

\subsection{DNN Encoder}

The DNN encoder block draws inspiration from the VGG architecture \cite{chang2020deepfake}, which has been recently tailored for face verification in \cite{cheng2020facial}. This encoder is structured with three primary blocks, with each block comprising two convolutional layers and one max-pooling layer. Following these blocks, there are two Fully Connected (FC) layers, having 1024 and 300 neurons, respectively. The function of the DNN encoder is to encode the input face image, reducing it to a 300-dimensional vector.

\section{Experimental Setup and Dataset}

\subsection{Experimental Setup}

The training phase utilized the CelebA dataset \cite{liu2018large}, while the evaluation was conducted on the Labeled Faces in the Wild (LFW) dataset \cite{huang2008labeled}. The performance metrics assessed were EER and accuracy.

During the selection of positives and negatives, we divided the validation partition of the CelebA dataset into two equal parts, creating datasets denoted as A and B, as explained previously. The CNN encoder was identical for both networks. The threshold values for the selection of positive and negative samples were 0.3 and 0.1, respectively. The activation function used for both the DNN and fully connected layers was ReLU, while the final layer of the network employed sigmoid activation function. The network is trained for 300 epochs or until the error is no longer decreasing using a batch size of 64 images. It uses stochastic gradient descent, momentum of 0.91, weight decay of 0.00001 and a logarithmically decaying learning rate from $10^{-2}$ 
 to $10^{-8}$. 

The baseline system underwent training utilizing the entire validation partition of the CelebA dataset. Its architecture entails a DNN encoder followed by a classification layer. To ensure an equitable comparison, the architecture of the CNN encoder was precisely replicated from that of the siamese networks. The training process involved minimizing cross-entropy loss, employing the Adam optimizer with identical parameters as those in the siamese networks. During evaluation, face image embeddings were extracted from the CNN encoder of this network.

\subsection{Datasets}

The CelebA dataset \cite{liu2018large}, initially introduced by Liu et al., comprises a vast collection of over 200,000 face images showcasing 10,177 celebrities. This dataset incorporates various elements like pose variations and background clutter, offering a comprehensive representation of real-world scenarios. To facilitate model training and evaluation, the dataset is divided into distinct sets for training, validation, and testing purposes. In this work, we utilized the training, validation, and test segments of the CelebA dataset as our unlabeled dataset for training the autoencoder.

The Labeled Faces in the Wild dataset (LFW) \cite{huang2008labeled} is a compilation of 13,233 images, featuring 5,749 distinct individuals. For testing, the dataset is randomly and uniformly divided into ten subsets. Each subset consists of 300 matched pairs (depicting the same person) and 300 mismatched pairs (depicting different individuals), which results in a total of 3,000 matched pairs and 3,000 mismatched pairs used for testing \cite{huang2008labeled}.

\begin{table}[]
\centering
\caption{It presents a comparison of the proposed method with different values of 
k and the baseline methods. The evaluation metrics used is Equal Error Rate (EER). The results demonstrate the performance of the proposed method across various k values, showcasing its effectiveness compared to the baseline methods.}
\label{table:experimental_result_baseline}
\begin{tabular}{lll}
\hline
\thead{Model} & \thead{k} &  \thead{EER} \\ \hline
Baseline         & -                        & 7.93        \\ \hline

Proposed      & 2                  & 7.91                        \\ \hline

Proposed               &  5                           & 7.83                         \\ \hline

Proposed  & 10 & 6.90                                     \\ \hline

Proposed               &  12                           & 7.01     \\ \hline

Proposed               & 15                            & 8.02                         \\ \hline

\end{tabular}
\end{table}

\section{Results}
\subsection{Results on LFW}

In Table \ref{table:experimental_result_baseline}, we present a comprehensive comparison between our proposed systems and the baseline, evaluating them based on the Equal Error Rate (EER) expressed as a percentage. The table encompasses results for different values of 
$k$. Throughout the experimental trials, face recognition scores are extracted from the output of the proposed network. The findings reveal that increasing the value of $k$ leads to enhanced system performance. The optimal EER of was achieved when $k$ was set to 10. While higher values of $k$ were explored, no significant improvement has been observed, suggesting a balance between performance and computational cost. This underscores the promising potential of the proposed approach, particularly in unsupervised scenarios, demonstrating its effectiveness in face verification tasks without the reliance on labeled training data.

Table \ref{table:experimental_result_LFW} provides a comprehensive comparison between our proposed method and state-of-the-art approaches. It's important to note that our method is compared with both supervised and unsupervised training techniques. Notably, our proposed training method demonstrates competitive performance without the need for an extensive labeled dataset. While some methods like ArcFace, GroupFace, Marginal Loss, and CosFace exhibit slightly higher accuracy, it's worth highlighting that our approach is trained on a significantly smaller dataset (approximately 200K images), whereas most state-of-the-art methods rely on millions of training images. This showcases the efficiency and effectiveness of our proposed method, especially in scenarios where labeled data is limited.

\begin{table}[]
\caption{Comparison of the proposed method result with some of state-of-the-art methods on LFW testing dataset.}
\label{table:experimental_result_LFW}
\begin{tabular}{lllll}
\hline
\thead{Model} & \thead{Training \\ size} &  \thead{Labeled/ \\ Unlabeled} &  \thead{Testing \\ size} & \thead{Accuracy(\%)} \\ \hline

UniformFace      & 6.1M                  & Labeled    & 6K                                    &  99.80 \cite{duan2019uniformface}                      \\ \hline

ArcFace               &  5.8M                           & Labeled   & 6K                                              & 99.82 \cite{deng2019arcface}                        \\ \hline

GroupFace  & 5.8M & labeled & 6K   & 99.85 \cite{kim2020groupface}                                    \\ \hline

CosFace               &  5M                           & Labeled    & 6K                                             & 99.73 \cite{wang2018cosface}                        \\ \hline

Marginal Loss               &  4M                           & Labeled   & 6K                                             & 99.48 \cite{deng2017marginal}                        \\ \hline

CurricularFace               &  3.8M                           & Labeled   & 6K                                             & 99.80 \cite{huang2020curricularface}                        \\ \hline

RegularFace               &  3.1M                           & Labeled    & 6K                                            & 99.61 \cite{zhao2019regularface}                        \\ \hline

MDCNN               &  1M                           & Labeled    & 6K                                            & 99.38 \cite{huang2022face}                        \\ \hline
PSO AlexNet TL               &  14M                           & Labeled     & 6K                                           & 99.57 \cite{elaggoune2022hybrid}                       \\
  
 \hline
Ben Face               &  0.5M                          & Labeled    & 6K                                            & 99.20 \cite{ben2021face}                   \\
 
 \hline

 F$^2$C               &  5.8M                          & Labeled   & 6K                                             & 99.83 \cite{wang2022efficient}                       \\
 \hline
PCCycleGAN               &  0.5M                          & Unlabeled    & 6K                                            & 99.52 \cite{liu2021unsupervised}                        \\
 
 \hline
CAPG GAN               &  1M                          & Unlabeled    & 6K                                            & 99.37 \cite{hu2018pose}                        \\

 \hline
 UFace               &  200K                          & Unlabeled    & 6K                                            & 99.40 \cite{solomon2022uface}                        \\

 \hline
 
\textbf{Proposed}      & \textbf{200K}                             & \textbf{Unlabeled}    & \textbf{6K}                                            & \textbf{99.67}              \\ \hline

\end{tabular}
\end{table}

\section{Conclusion}

In this work, we introduced a siamese network for face verification that operates without the reliance on labeled data. This network features two branches, each functioning as a CNN encoder, and was trained as a binary classifier. Given our objective of avoiding the need for face image labels, the training sample pairs were generated using an unsupervised approach. Positive training images were selected within a single dataset based on the highest cosine scores with a designated anchor, while negative training images followed the same selection criteria but were drawn from a different dataset.

Following the training process, decision scores were obtained using the proposed network, and the evaluation was conducted using the LFW dataset. Notably, our experimental results demonstrated that our proposed system, despite being unsupervised, achieved results that closely paralleled those of fully supervised baselines. This highlights the effectiveness of our approach in face verification, even in the absence of labeled data.

\end{document}